\documentclass[10pt,twocolumn,letterpaper]{article}

\usepackage[pagenumbers]{cvpr} 

\usepackage[table, dvipsnames]{xcolor}

\usepackage{multirow}
\usepackage{indentfirst}

\definecolor{cvprblue}{rgb}{0.21,0.49,0.74}
\usepackage[pagebackref,breaklinks,colorlinks,citecolor=cvprblue]{hyperref}

\def\paperID{} 
\def\confName{}
\def\confYear{}

\title{Distilling Temporal Knowledge with Masked Feature Reconstruction for 3D Object Detection}

\author{Haowen Zheng$^1$, Dong Cao$^2$, Jintao Xu$^2$, Rui Ai$^2$, Weihao Gu$^2$, Yang Yang$^3$, Yanyan Liang$^1$ \\
{$^1$}Macau University of Science and Technology \\
{$^2$}HAOMO.AI Technology Co., Ltd.\\
{$^3$}State Key Laboratory of Multimodal Artificial Intelligence Systems,\\ 
	Institute of Automation, Chinese Academy of Sciences
}

\begin{document}
\maketitle
\begin{abstract}
Striking a balance between precision and efficiency presents a prominent challenge in the bird’s-eye-view (BEV) 3D object detection. Although previous camera-based BEV methods achieved remarkable performance by incorporating long-term temporal information, most of them still face the problem of low efficiency. One potential solution is knowledge distillation. Existing distillation methods only focus on reconstructing spatial features, while overlooking temporal knowledge. To this end, we propose \textbf{TempDistiller}, a \textbf{Temp}oral knowledge \textbf{Distiller}, to acquire long-term memory from a teacher detector when provided with a limited number of frames. Specifically, a reconstruction target is formulated by integrating long-term temporal knowledge through self-attention operation applied to the feature of teachers. Subsequently, novel features are generated for masked student features via a generator. Ultimately, we utilize this reconstruction target to reconstruct the student features. In addition, we also explore temporal relational knowledge when inputting full frames for the student model. We verify the effectiveness of the proposed method on the nuScenes benchmark. The experimental results show our method obtain an enhancement of +1.6 mAP and +1.1 NDS compared to the baseline, a speed improvement of approximately 6 FPS after compressing temporal knowledge, and the most accurate velocity estimation.
\end{abstract}    
\section{Introduction}
\label{sec:intro}

\begin{figure}[t]
\begin{center}
\includegraphics[width=\linewidth]{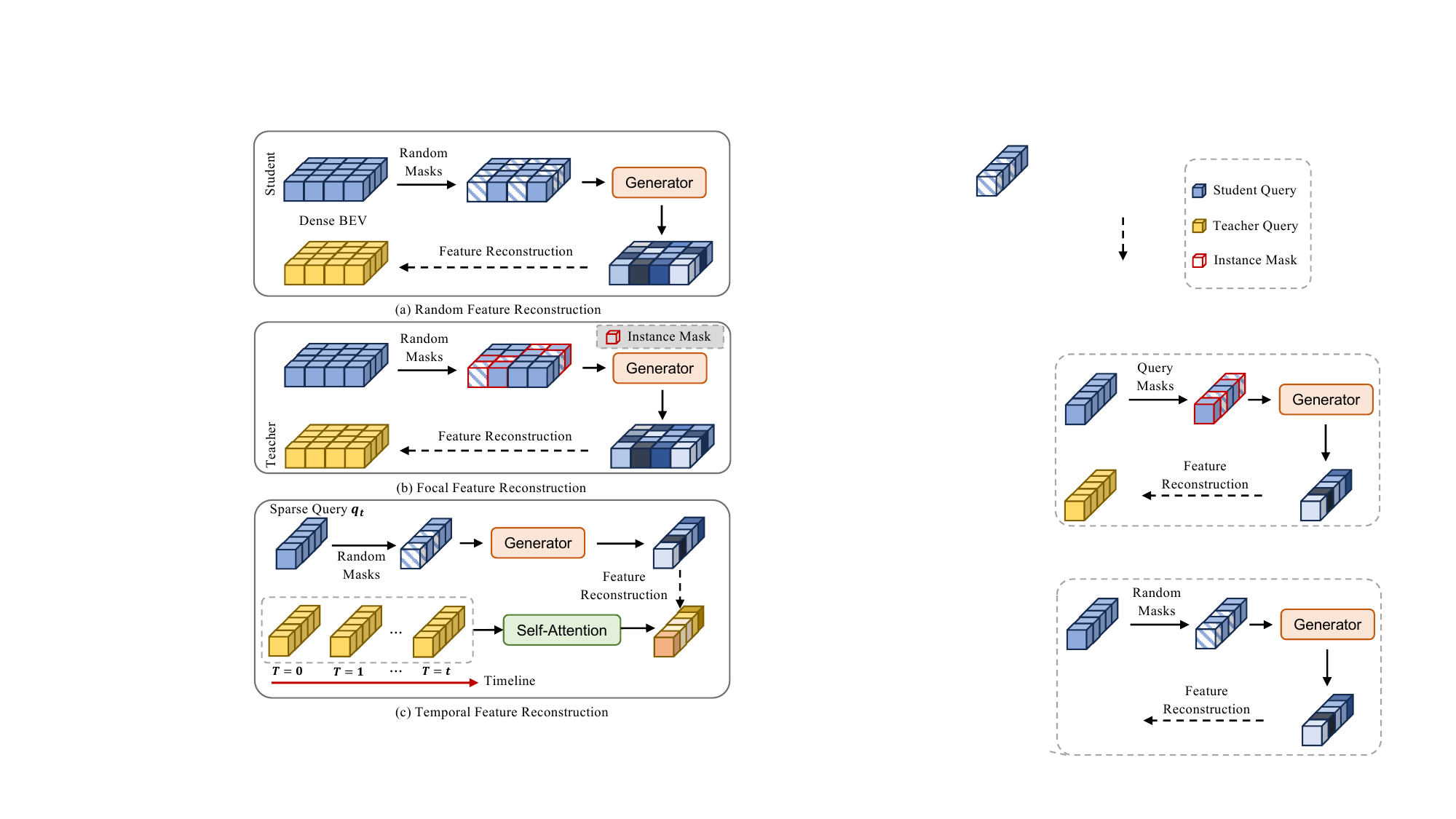}
\end{center}
\vspace{-0.8em}
\caption{Overview of masked feature reconstruction. (a) Random masks are generated on student features. Then spatial features are recovered from the teacher. (b) Instance masks are introduced to filter foreground areas. Random masks are then generated within these specific areas and student features are reconstructed from the teacher. (c) The proposed temporal feature reconstruction is based on sparse BEV representation. The reconstruction target involves temporal aggregated features derived from a teacher model, facilitating the acquisition of long-term temporal knowledge by student features.}
\vspace{-1.0em}
\label{fig:comparison}
\end{figure}

Camera-based 3D object detection has attracted much attention in autonomous driving due to its low deployment cost and rich visual information. Recently, bird’s-eye-view (BEV) detection achieves promising performance by incorporating temporal information \cite{li2022bevformer, yang2023bevformer, huang2022bevdet4d, li2023bevdepth, liu2023petrv2, park2022time, han2023exploring, lin2023sparse4d, liu2023sparsebev, wang2023exploring, lin2022sparse4d}. However, achieving a balance between accuracy and efficiency remains a significant challenge in BEV detection. One potential solution involves employing knowledge distillation to enhance the performance of compact detectors. General 2D distillation techniques \cite{shu2021channel, yang2022masked} can be easily adapted to 3D detection. Specifically tailored for 3D detection, FD3D \cite{zeng2023distilling} first presents a focal distiller to capture local knowledge within foreground areas. Distilling in the camera-only setting avoids modality alignment and heterogeneous problems from different modalities, which is caused by LiDAR-based knowledge distillation \cite{zhou2023unidistill, wang2023distillbev, liu2023geomim, li2022unifying}.

Although these distillation methods and temporal fusion methods are effective, we discover they are still problematic in three aspects. (i) The distillation methods \cite{shu2021channel, yang2022masked, zeng2023distilling} primarily concentrate on spatial information while neglecting temporal knowledge. (ii) \cite{huang2022bevdet4d, li2023bevdepth, liu2023petrv2} lack the ability for long-term memory and \cite{lin2022sparse4d, liu2023sparsebev, park2022time} increase computational complexity when handling multiple frames concurrently in a parallel temporal fusion paradigm. To alleviate the issues, \cite{lin2023sparse4d, han2023exploring, wang2023exploring} aggregate temporal features in a sequential paradigm. However, we argue that such methods may suffer from temporal knowledge forgetting. (iii) Dense BEV representation \cite{li2022bevformer, yang2023bevformer, li2023bevdepth, huang2022bevdet4d, huang2021bevdet, park2022time} leads to high computation overhead, while sparse BEV representation \cite{lin2022sparse4d, liu2023petrv2, liu2023sparsebev} is still affected by the aforementioned second problem.

Given the aforementioned problems, we provide a novel perspective to process temporal information in sparse BEV space through knowledge distillation. Our proposed method, termed TempDistiller, is a temporal knowledge distiller based on masked feature reconstruction. The core idea of masked feature reconstruction refers to leveraging partial pixels from a student model to reconstruct the complete features, guided by a teacher model. Such a reconstruction paradigm aims to help the student to achieve better representation. As illustrated in Fig. \ref{fig:comparison} (a) and (b), previous feature reconstruction methods \cite{yang2022masked, zeng2023distilling, huang2022masked} only recover spatial features. Different from them, we focus on reconstructing temporal knowledge for student features. In Fig. \ref{fig:comparison} (c), we construct a reconstruction objective by aggregating long-term temporal knowledge through a self-attention operation applied to teacher features. Then we introduce random masks to student features and generate new features by a generator. Finally, we reconstruct the student features based on the reconstruction objective. This design enables the student detector to assimilate long-term temporal knowledge without causing additional computational overhead during inference.

In addition, we empirically observe that when the number of input frames for the student matches that of the teacher, the effect of temporal feature reconstruction is minimal. Given that the student model inherently learns long-term temporal information from multiple frames, the knowledge provided by temporal feature reconstruction reaches a point of saturation. In response, we shift our focus to exploring temporal relational distillation as a complementary technique. This method is adept at capturing the similarity of objects across frames, particularly with respect to moving objects, thereby playing a crucial role in enhancing velocity estimation.

In summary, our contributions can be described as
\begin{itemize}
\item To the best of our knowledge, this is the first work to transfer temporal knowledge on 3D object detection. We explore how to learn temporal knowledge and its relational information. 
\item We provide a novel view to process long-term temporal fusion with knowledge distillation. Even with a reduced number of input frames, our method allows a student detector to glean long-term temporal knowledge from a teacher detector through temporal feature reconstruction. 
\item The proposed method shows a performance improvement of 1.6 mAP and 1.1 NDS compared to the baseline and an inference speed improvement of about 6 FPS after reducing the number of input frames. Moreover, we achieve the most accurate velocity estimation. From the qualitative results, we also improve the detection of occluded and distant objects.
\end{itemize}

\section{Related Work}
\label{sec:related_work}

\subsection{Surround-view 3D Object Detection in BEV}
Recently, surround-view camera-based 3D object detection has achieved significant success with bird's eye view (BEV) representation. For instance, Lift-Splat-Shoot (LSS) \cite{philion2020lift} transforms multi-view 2D image features into 3D space based on depth estimation. Building upon LSS, BEVDet series \cite{huang2021bevdet, huang2022bevdet4d, li2023bevdepth} further elevate performance by introducing data augmentation, temporal fusion, and LiDAR-derived depth ground truth. Different from 2D-to-3D back-projection methods, BEVFormer \cite{li2022bevformer} constructs a dense BEV space to sample multi-view 2D features using a deformable attention mechanism. Inspired by DETR \cite{carion2020end}, DETR3D \cite{wang2022detr3d} initializes a set of 3D queries instead of dense BEV queries to explore a sparse BEV representation. However, the performance is compromised if projecting 3D query points to 2D space for sampling camera features with a fixed local receptive field. To address this issue, PETR series \cite{liu2022petr, liu2023petrv2} leverage global attention to expand the receptive field. Despite the remarkable progress achieved by the aforementioned approaches, they still entail substantial computational burdens. Consequently, SparseBEV \cite{liu2023sparsebev} presents a fully sparse detector with a scale-adaptive receptive field for a better trade-off between accuracy and speed. Nevertheless, SparseBEV still suffers from slowdowns when processing longer frames. To handle this issue, an intuitive way is to harness knowledge distillation (KD). In this paper, we delve into temporal knowledge within sparse BEV representation.

\begin{figure*}[ht]
\begin{center}
\includegraphics[width=\linewidth]{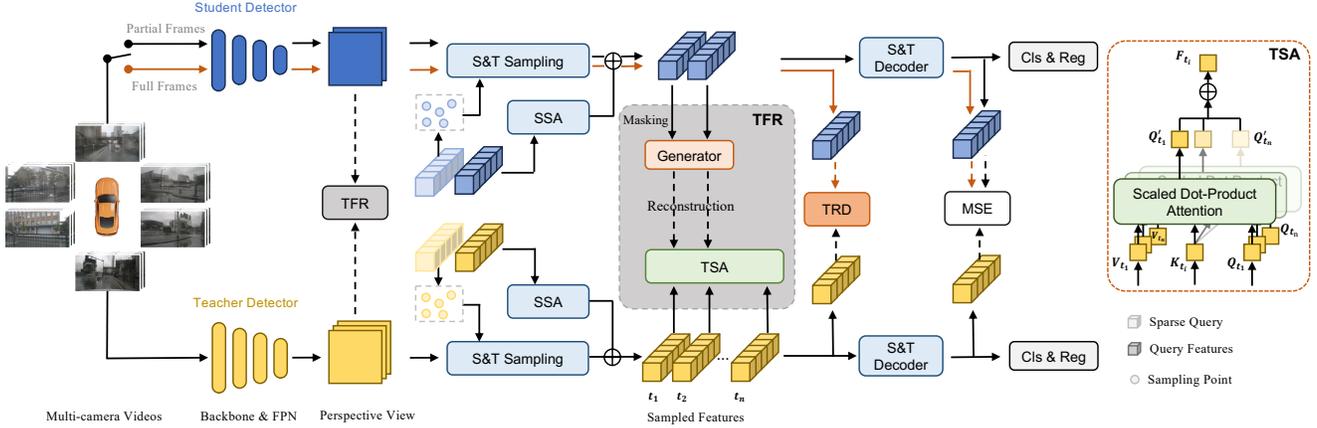}
\end{center}
\caption{Overall framework of TempDistiller. The proposed method aims to enable student detector to learn long-term temporal knowledge from teacher detector, particular with fewer input frames. Taking the long-term temporal teacher features as the reconstruction objective, we leverage temporal feature reconstruction (TFR) to force student detector to study enhanced representation of perspective features and BEV features. Additionally, we explore temporal relational distillation (TRD) when the student detector is fed with full frames. Finally, we impose constraints on the features after S\&T (Spatio-Temporal) decoder by MSE (mean square error), which encourages the learning of semantically rich spatio-temporal features. SSA refers to the spatial self-attention operation in \cite{liu2023sparsebev}.}
\label{fig:framework}
\end{figure*}

\subsection{Temporal Modeling}
Integrating historical information, especially long-term temporal knowledge is crucial in autonomous driving. Mainstream temporal fusion mechanisms can be divided into parallel temporal fusion \cite{yang2023bevformer, liu2023petrv2, huang2022bevdet4d, li2023bevdepth, park2022time, liu2023sparsebev} and sequential temporal fusion \cite{li2022bevformer, wang2023exploring, han2023exploring, lin2023sparse4d}. Early works \cite{liu2023petrv2, huang2022bevdet4d, li2023bevdepth} fuse short-term memory (2-4 frames), but their performance is not satisfactory. To explore long-term temporal fusion, SOLOFusion \cite{park2022time} utilizes 17 frames and achieves outstanding performance. However, such parallel temporal fusion methods commonly grapple with the challenge of balancing accuracy and efficiency. To alleviate this problem, \cite{wang2023exploring, han2023exploring, lin2023sparse4d} carry out sequential temporal fusion instead. They propagate historical features into the current timestamp, largely accelerating the inference speed while maintaining excellent performance. In this paper, we focus on the parallel temporal fusion paradigm and approach the problem from a novel perspective by compressing temporal information through knowledge distillation.

\subsection{Knowledge Distillation for Object Detection}
Applying knowledge distillation \cite{hinton2015distilling} on 2D object detection is a popular topic. Different from distilling global features \cite{chen2017learning}, several works emphasize the importance of region selection based on bounding boxes \cite{wang2019distilling, dai2021general, guo2021distilling} and the application of attentive masks on features \cite{zhang2020improve, huang2022masked} to mitigate noise interference. FGD \cite{yang2022focal} effectively combines both strategies, resulting in further performance enhancements. From another perspective, masked feature reconstruction has proved its effectiveness. Inspired by masked image modeling, MGD \cite{yang2022masked} generates random masks on student features, and then reconstructs them under the guidance of teacher features. Huang \etal \cite{huang2022masked} generate attentive masks instead of random masks for feature reconstruction.

For 3D object detection, most methods \cite{zhou2023unidistill, wang2023distillbev, liu2023geomim, li2022unifying} focus on cross-modality knowledge distillation, aiming to transfer LiDAR-based features to camera-based features. Although these methods notably improve performance by leveraging LiDAR's real world modeling capabilities, aligning different modality poses a significant challenge, leading to difficulties to process heterogeneous problem. To this end, FD3D \cite{zeng2023distilling} first proposes camera-only distillation to reconstruct focal knowledge from imperfect teachers. We also follow the setting of camera-only distillation in this study. Previous masked feature construction methods only concern about spatial knowledge, neglecting temporal knowledge modeling. In this work, we undertake the reconstruction of long-term temporal knowledge for the student model, guided by the teacher model. This approach empowers the student model with long-term memory even with a limited number of input frames.

\section{Method}
Fig. \ref{fig:framework} demonstrates the pipeline of TempDistiller. We first describe the overall framework in Sec. \ref{sec:framework}. In Sec. \ref{sec:SR}, we depict an efficient query-based method for sparse BEV representation. Then a temporal feature reconstruction method is elaborated in Sec. \ref{sec:TFR}. We further investigate temporal relational knowledge in Sec. \ref{sec:TRD}. Ultimately, we introduce the overall loss in Sec. \ref{ovelall_loss}.

\subsection{Overall Framework}
\label{sec:framework}
Incorporating multiple frames \cite{huang2022bevdet4d, liu2023petrv2, park2022time, liu2023sparsebev} can offer richer motion information, which greatly improves the detection and velocity estimation performance for dynamic objects. However, within the parallel temporal fusion paradigm, we observe that an increasing in the number of input frames leads to a reduction in inference speed, whereas too few input frames result in decreased precision. Consequently, striking a balance between accuracy and speed becomes challenging.

To tackle this challenge, we introduce TempDistiller, as illustrated in Fig. \ref{fig:framework}, aiming to strike a better balance between precision and efficiency. The core components of TempDistiller encompass the temporal feature reconstruction loss and temporal relational distillation loss. Benefiting from them, one essential characteristic of TempDistiller lies in its ability to capture long-term temporal memory even with a limited number of input frames. Another notable attribute is the capacity to extract inter-frame relations. In our approach, we utilize a heavy BEV detector as the teacher model and transfer temporal knowledge to a lightweight student BEV detector. The teacher detector can process a long sequence of frames, while the input frame length for the student model is optional. During the distillation stage, we freeze the teacher detector and retain the original student architecture by removing any extra auxiliary layers post-training. Based on this, there is no additional computational overhead during the inference. 

\subsection{Sparse BEV Representation}
\label{sec:SR}
Dense BEV representation introduces computational complexity, despite attempts to alleviate it by leveraging knowledge distillation on lightweight detectors \cite{yang2022focal}. Therefore, we turn to sparse BEV representation, following the approach outlined in \cite{wang2022detr3d, liu2023sparsebev}. In this representation, a sparse query is defined as a nine-tuple $Q=(x, y, z, w, l, h, \theta, v_x, v_y)$, where $(x, y, z)$ denotes the query's coordinate in the BEV space, while $w, l, h$ represent its width, length, and height, respectively. $\theta$ and $(v_x, v_y)$ indicate query's rotation and velocity. The query set consists of $N_q$ queries, each associated with $C$-dim features. Initially, these query features are transmitted to the spatial self-attention with a scalable receptive field \cite{liu2023sparsebev}. Subsequently, a set of 3D points generated by these queries are projected onto the FPN feature maps to sample spatial features. The resulting sampled features $\textbf{F} \in \mathbb{R}^{T \times N_q \times C}$ aggregate multi-view and multi-scale information for each frame, where T denotes the number of frames. Note that before sampling, the sparse query points are aligned to the features at the current timestamp based on the ego-motion and velocity of other dynamic objects. Following this, the sampled features are added to the query features and proceed through a decoder for adaptive spatio-temporal feature decoding \cite{liu2023sparsebev}. In our temporal knowledge distillation, we perform temporal feature reconstruction on both the sampled features $\textbf{F}$ and FPN features. Additionally, we explore inter-frame relations via temporal relational distillation. Finally, we conduct feature alignment on the decoded features.

\subsection{Temporal Feature Reconstruction}
\label{sec:TFR}
The sampled features rely on input from multiple frames, necessitating the backbone and BEV encoder/decoder to handle long temporal sequences. Disparate from it, we provide a novel view to transfer long-term memory via reconstructing temporal knowledge. Initially, we randomly mask a proportion of the student features. Then the student features are recovered under the guidance of the teacher features. To facilitate the acquisition of long-term temporal knowledge by the student features, we aggregate the teacher features with more frames through self-attention operations.

Specifically, we denote the student features and the teacher features as $\textbf{F}_{stu} \in \mathbb{R}^{T_{stu} \times N_q \times C}$ and $\textbf{F}_{tea} \in \mathbb{R}^{T_{tea} \times N_q \times C}$, respectively, where $T_{tea}=T_{stu}+k$, $8 > k \geq 0$. Random mask is generated on $\textbf{F}_{stu}$ with mask ratio $\lambda$, which can be formulated as 

\begin{equation}
    M_{i,t} = \begin{cases}
    0, \text{if} \ R_{i,t} < \lambda \\
    1, \text{otherwise}
    \end{cases},
\end{equation}
where $R_{i,t}$ is a random number in $(0,1)$ and $i$ indicates the index of query. $t$ denotes the $t$-th frame. Moreover, we perform temporal self-attention (TSA) to aggregate teacher temporal information. The sampled teacher features from all frames serve as queries and values. The features from the first $T_{stu}$ frames are represented as keys, which align the temporal sequence length with that of the student features. Therefore, the aggregation of teacher temporal knowledge can be formulated as

\begin{equation}
    \textbf{F}_{agg}^{t} = \sum_{t_1=1}^{t+k} \text{TSA}(\textbf{F}_{tea}^{t_1}, \textbf{F}_{tea}^{t}, \textbf{F}_{tea}^{t_1}),
    \label{eq2}
\end{equation}
where TSA is based on the scaled dot-product attention operation \cite{vaswani2017attention}.

Once the reconstruction objective is established, we then recover masked student features using a generation layer $\mathcal{G}$:

\begin{equation}
    \hat{\textbf{F}}_{stu} = \mathcal{G}(\textbf{F}_{stu} \cdot M),
    \label{eq3}
\end{equation}
where $\mathcal{G}$ is composed of two 1D $3 \times 3$ convolutional layers and one $ReLU$ layer. Then the generated student features $\hat{\textbf{F}}_{stu}$ are reconstructed under the supervision of the teacher temporally aggregated features $\textbf{F}_{agg}^{t}$ using Mean Squared Error (MSE). Thus the temporal reconstruction loss for sparse BEV features can be represented as

\begin{equation}
    \mathcal{L}_{rc}^{bev} = \frac{1}{N} \sum_{t=1}^{T_{stu}} \sum_{q=1}^{N_q} \sum_{c=1}^{C} \Vert \hat{\textbf{F}}_{stu}^{t,q,c} - \textbf{F}_{agg}^{t,q,c}\Vert^2_2,
    \label{eq4}
\end{equation}
where $N=T_{stu} \times N_q \times C$.

Moreover, the temporal feature reconstruction method is not limited to BEV and can seamlessly extend to the perspective view (PV). To achieve this, we apply the temporal feature reconstruction technique to the final layer features of the FPN. As the spatial shape of feature maps in PV is two-dimensional, the convolutional layers within $\mathcal{G}$ are modified to 2D convolution. Similarly, the reconstruction loss for PV can be formulated as

\begin{equation}
    \mathcal{L}_{rc}^{pv} = \frac{1}{N} \sum_{t=1}^{T_{stu}} \sum_{l=1}^{L} \sum_{c=1}^{C} \Vert \hat{\textbf{F}}_{stu}^{t,l,c} - \textbf{F}_{agg}^{t,l,c}\Vert^2_2,
    \label{eq5}
\end{equation}
where $L=H \times W$ and $N=T_{stu} \times L \times C$. H and W are the height and width of FPN features, respectively. For the first three layers of the FPN, we employ spatial feature reconstruction method like MGD \cite{yang2022masked}.

In our experiments, we surprisingly find that reconstructing temporal features from a smaller teacher detector (utilizing the ResNet50 backbone) can achieve comparable performance compared to that of a larger teacher detector (employing the ResNet101 backbone). It precisely reveals the effectiveness of our method in learning long-term temporal knowledge, which does not rely on a stronger backbone.

\begin{figure}[t]
\begin{center}
\includegraphics[width=\linewidth]{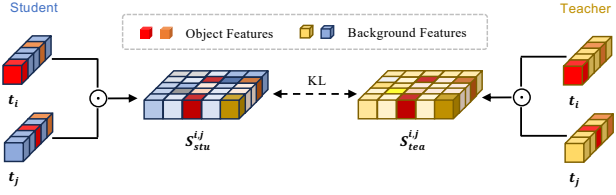}
\end{center}
\caption{Overview of the proposed temporal relational distillation (TRD). $\odot$ indicates element-wise product. Objects that are related in different frames have high responses in the similarity matrix.}
\label{fig:trd}
\end{figure}

\subsection{Temporal Relational Distillation}
\label{sec:TRD}
When the number of input frames of the student model is equal to that of the teacher model, we observe that the effectiveness of temporal feature reconstruction diminishes. This occurrence might stem from the inherent capability of the student model to capture long-term temporal information adequately with the available frames, rendering temporal feature reconstruction less impactful. As an alternative, we propose temporal relational distillation for this case (see Fig. \ref{fig:trd}). This method computes feature similarity between each frame to represent relational knowledge. Specifically, given sparse query features $\textbf{F}_{i}$ and $\textbf{F}_j$ from the $i$-th frame and the $j$-th frame, the feature similarity is represented as $S^{i,j}=\textbf{F}_{i}\textbf{F}_{j}^\top \in \mathbb{R}^{N \times C}$, where $N=N_q \times N_q$. It is particularly useful for dynamic object detection, as moving objects tend to show high responses in the similarity matrix. We transfer such relational knowledge from the teacher model to the student model using KL-divergence:

\begin{equation}
    \mathcal{L}_{trd} = \frac{1}{N} \sum_{n=1}^{N} KL(\sigma(\frac{S^{i,j,n}_{stu}}{\tau}) \Vert \sigma(\frac{S^{i,j,n}_{tea}}{\tau})),
    \label{eq6}
\end{equation}
where $\sigma$ is softmax and $\tau$ is a temperature (set to 0.5).

\subsection{Overall Distillation Loss}
\label{ovelall_loss}
In addition to the aforementioned losses, we further constrain the decoded features $D \in \mathbb{R}^{N_q \times C}$ by L2 loss:

\begin{equation}
    \mathcal{L}_{dc} = \frac{1}{N} \sum_{q=1}^{N_q} \sum_{c=1}^{C} \Vert D_{stu}^{q,c} - D_{tea}^{q,c} \Vert^2_2,
    \label{eq7}
\end{equation}
where $N=N_q \times C$. The decoded features encapsulate high-level spatio-temporal information, which is beneficial for the student model to learn abstract knowledge. Therefore, combining Eq. \ref{eq4}, Eq. \ref{eq5}, Eq. \ref{eq6} and Eq. \ref{eq7}, the overall distillation loss is formulated as

\begin{equation}
    \mathcal{L}_{dist} = \alpha_1 \mathcal{L}_{rc}^{bev} + \alpha_2 \mathcal{L}_{rc}^{pv} + \alpha_3 \mathcal{L}_{dc} + \alpha_4 \mathcal{L}_{trd},
    \label{eq8}
\end{equation}
where $\alpha_1$, $\alpha_2$, $\alpha_3$ and $\alpha_4$ are loss weights to balance the losses. $\alpha_4$ is set to 0 when inputting partial frames for the student model, while for inputting full frames, $\alpha_1$ and $\alpha_2$ are set to 0. In summary, we train the student model with original classification and regression loss as well as the overall distillation loss.
\section{Experiments}

\begin{table*}[t]

\begin{center}
\resizebox{\linewidth}{!}{
\begin{tabular}{l|c|c|cc|ccccc|c}
\toprule[1pt]
Method         & Backbone & Frames & NDS $\uparrow$ & mAP $\uparrow$  & mATE $\downarrow$ & mASE $\downarrow$ & mAOE $\downarrow$ & mAVE $\downarrow$ & mAAE $\downarrow$ & FPS $\uparrow$ \\  \midrule
BEVDet $\ddag$ \cite{huang2021bevdet} & ResNet50 & 1 & 37.9 & 29.8 & 0.725 & 0.279 & 0.589 & 0.860 & 0.245 & - \\
BEVDet4D $\ddag$ \cite{huang2022bevdet4d} & ResNet50 & 2 & 45.7 & 32.2 & 0.703 & 0.278 & 0.495 & 0.354 & 0.206 & 30.7 \\
PETRv2 \cite{liu2023petrv2} & ResNet50 & 2 & 45.6 & 34.9 & 0.700 & 0.275 & 0.580 & 0.437 & 0.187 & - \\
SOLOFusion $\ddag$ \cite{park2022time} & ResNet50 & 16+1 & 53.4 & 42.7 & 0.567 & 0.274 & 0.511 & 0.252 & 0.181 & 15.7* \\ \midrule
VideoBEV $\ddag$ \cite{han2023exploring} & ResNet50 & 8 & 53.5 & 42.2 & \textbf{0.564} & 0.276 & 0.440 & 0.286 & 0.198 & - \\
Sparse4Dv2 \cite{lin2023sparse4d}  & ResNet50 & - & 53.9 & 43.9 & 0.598 & 0.270 & 0.475 & 0.282 & \textbf{0.179} & 17.3 \\
StreamPETR $\dag$ \cite{wang2023exploring}  & ResNet50 & 8 & 55.0 & 45.0 & 0.613 & \textbf{0.267} & 0.413 & 0.265 & 0.196 & \textbf{33.9} \\ \midrule
\multicolumn{11}{c}{Results w/ and w/o distillation schemes} \\ \midrule
T:BEVFormer \cite{li2022bevformer} & ResNet101-DCN & 4      & 51.7 & 41.6 & 0.673 & 0.274 & 0.372 & 0.394 & 0.198 & 4.3 \\ 
S:BEVFormer & ResNet50 & 3      & 46.3 & 34.6 & 0.743 & 0.280 & 0.445 & 0.440 & 0.194 & 13.7 \\
+FD3D \cite{zeng2023distilling} & ResNet50 & 3      & 48.7 & 37.6 & 0.719 & 0.275 & 0.409 & 0.394 & 0.212 & 13.7 \\ \midrule
T:SparseBEV $\dag$ \cite{liu2023sparsebev} & ResNet101 & 8      & 59.2 & 50.1 & 0.562 & 0.265 & 0.321 & 0.243 & 0.195 & 7.2 \\
S:SparseBEV $\dag$ & ResNet50  & 4      & 52.9 & 41.6 & 0.630 & 0.270 & 0.435 & 0.263 & 0.187 & 26.1 \\ 
+MGD \cite{yang2022masked}  & ResNet50 & 4      & 53.3 & 42.3 & 0.636 & 0.270 & 0.440 & 0.258 & 0.184 & 26.1 \\
+CWD \cite{shu2021channel} & ResNet50 & 4      & 53.4 & 42.4 & 0.624 & 0.274 & 0.448 & 0.255 & 0.183 & 26.1 \\
\rowcolor{gray!40}
+TempDistiller  & ResNet50 & 4      & 54.0 & 43.2 & 0.620 & 0.272 & 0.429 & 0.253 & 0.181 & 26.1 \\ 
S:SparseBEV $\dag$ & ResNet50  & 8      & \textbf{55.5} & 44.7 & 0.585 & 0.271 & \textbf{0.391} & 0.251 & 0.188 & 20.2 \\
\rowcolor{gray!40}
+TempDistiller  & ResNet50 & 8      & \textbf{55.5} & \textbf{45.1} & 0.591 & \textbf{0.267} & 0.421 & \textbf{0.238} & 0.184 & 20.2 \\ \midrule

T:SparseBEV $\dag$ & ResNet50  & 8      & 55.5 & 44.7 & 0.585 & 0.271 & 0.391 & 0.251 & 0.188 & 20.2 \\
S:SparseBEV $\dag$ & ResNet50  & 4      & 52.9 & 41.6 & 0.630 & 0.270 & 0.435 & 0.263 & 0.187 & 26.1 \\ 
\rowcolor{gray!40}
+TempDistiller  & ResNet50 & 4      & 54.0 & 42.8 & 0.619 & 0.270 & 0.395 & 0.259 & 0.198 & 26.1 \\

\bottomrule[1pt]
\end{tabular}
}
\end{center}
\vspace{-0.8em}
\caption{Comparison on the nuScenes validation set. $\dag$ indicates benefits from perspective pretraining. $\ddag$ denotes methods with CBGS. FPS is measured on RTX4090 with fp32 without cuda acceleration. * represents inference with fp16. The input size for ResNet101 and ResNet101-DCN is 900 $\times$ 1600, whereas for ResNet50 it is 256 $\times$ 704, except for \cite{zeng2023distilling} and its baseline (450 $\times$ 800).}
\vspace{-0.8em}
\label{tab1}
\end{table*}

\subsection{Datasets and Metrics}
A large-scale autonomous driving benchmark nuScenes \cite{caesar2020nuscenes} is utilized to evaluate our approach. It comprises 700/150/150 scenes for training/validation/testing. Each scene spans approximately 20 seconds, with annotations available for keyframes at 0.5s intervals. The dataset captures frames using six cameras, providing a 360-degree field of view. For 3D object detection, it includes 1.4M 3D bounding boxes across 10 categories. Following the official evaluation metrics, we report nuScenes detection score (NDS), mean Average Precision (mAP), and five true positive (TP) metrics including ATE, ASE, AOE, AVE, and AAE for measuring translation, scale, orientation, velocity and attributes respectively. The NDS combines mAP and five TP metrics to provide a comprehensive evaluation score.

\begin{table}[t]
\resizebox{\linewidth}{!}{
\begin{tabular}{ccc|cc|c}
\toprule[1pt]
\multicolumn{2}{c}{TFR} & \multirow{2}{*}{Decoded Features} & \multirow{2}{*}{NDS $\uparrow$} & \multirow{2}{*}{mAP $\uparrow$} & \multirow{2}{*}{mAVE $\downarrow$} \\ \cline{1-2}
PV   & BEV   & &  &  &  \\ \midrule
                             &       &                                  & 52.9 & 41.6 & 0.263 \\
\multicolumn{1}{c} {\checkmark} &       &                                  & 53.5 & 42.3 & 0.264 \\
                         & {\checkmark} &                                  & 53.5 & 42.7 & 0.255 \\ 
                         &  & {\checkmark} & 53.6 & 42.3 & 0.262 \\ 
                         & {\checkmark} & {\checkmark} & 53.3 & 42.5 & 0.253 \\ 
\multicolumn{1}{c} {\checkmark} & {\checkmark} &  & 53.7 & 43.0 &  \textbf{0.249} \\
\rowcolor{gray!40}
\multicolumn{1}{c} {\checkmark} & {\checkmark} & {\checkmark} & \textbf{54.0} & \textbf{43.2} & 0.253 \\
\bottomrule[1pt]
\end{tabular}
}
\caption{Effectiveness of loss components. TFR indicates temporal features reconstruction, which is applied on perspective view (PV) and bird’s-eye-view (BEV), respectively.}
\vspace{-1.0em}
\label{tab2}
\end{table}

\subsection{Implementation Details} 
Our method is implemented with PyTorch. The experimental results are reported based on 8 A100 GPUs and FPS measurements are conducted on RTX4090 with fp32. We train all the models with AdamW \cite{loshchilov2017decoupled} optimizer for 24 epochs, using perspective pretraining on nuImage \cite{caesar2020nuscenes}. The initial learning rate is set to $2 \times 10^{-4}$ and is decayed with cosine annealing policy. The global batch size is fixed to 8. For supervised training, the Hungarian algorithm \cite{kuhn1955hungarian} is used for label assignment. Focal loss \cite{lin2017focal} and L1 loss are employed for classification and 3D bounding boxes regression, respectively. We process adjacent frames at intervals of 0.5 seconds.

For modeling temporal knowledge with sparse BEV representation, we adopt the ResNet101 and ResNet50 backbone \cite{he2016deep} from SparseBEV \cite{liu2023sparsebev}, benefiting from their publicly available model weights. The choice of the teacher model includes ResNet101 or ResNet50, while the student model utilizes ResNet50 with the flexibility of adjusting number of input frames. The input size is set to 900 $\times$ 1600 for ResNet101 and it 256 $\times$ 704 for ResNet50. We initialize $N_q=900$ queries and set the channel of query features $C=256$. The mask ratio $\lambda$ is fixed to 0.5. Loss weights $\alpha_1$, $\alpha_2$, $\alpha_3$ and $\alpha_4$ are set to $5e^{-4}$, $1e^{-3}$, 1 and 1, respectively. In addition, query denoising \cite{li2022dn} is used for training stabilization and faster convergence following \cite{liu2023sparsebev, wang2023exploring}.

\subsection{Main Results} 
\textbf{Comparison with camera-only distillation methods.} FD3D \cite{zeng2023distilling} is the first distillation method that works on a camera-only setting, which is designed for spatial dimension. Directly extending its application to temporal features may cause a huge computational burden, especially considering that the generator needs to produce new features for each temporal dimension. Therefore, It's difficult to apply it to parallel temporal fusion methods \cite{liu2023sparsebev}. TempDistiller comprehensively outperforms FD3D based on a more powerful temporal detector. For a fairer comparison, we also explore other distillation methods \cite{yang2022masked, shu2021channel}. Our method surpasses MGD by 0.9 mAP and 0.7 NDS and exceeds CWD by 0.8 mAP and 0.6 NDS. Additionally, our method achieves the lowest mATE, mAOE, mAVE, and mAAE compared to these methods.

\textbf{Comparison with state-of-the-art methods.}
State-of-the-art (SOTA) temporal methods can be divided into parallel temporal fusion \cite{huang2022bevdet4d, liu2023petrv2, park2022time, liu2023sparsebev, lin2022sparse4d} and sequential temporal fusion \cite{lin2023sparse4d, han2023exploring, wang2023exploring}. Our method boosts 0.4 mAP and achieves the lowest mAVE of 0.238 compared with SparseBEV \cite{liu2023sparsebev} ($T_{stu}=8$). In terms of parallel paradigms, TempDistiller exhibits the best performance using fewer frames ($T_{stu}=4$), surpassing SOLOFusion \cite{park2022time} in both accuracy and inference speed. Compared to StreamPETR \cite{wang2023exploring}, we observe the 1.2\% and 1.5\% improvement in mAVE and mAAE, respectively, and a slight enhancement in mAP. In addition, we find an interesting result that leveraging ResNet50 as the backbone yields comparable performance to using a larger backbone (i.e., ResNet101). This result eliminates the impact of the difference in capabilities between backbones and solidly demonstrates that the proposed method can learn long-term temporal knowledge with fewer input frames.

\begin{table}[t]
\centering
\resizebox{\linewidth}{!}{
\begin{tabular}{c|c|cc|c}
\toprule[1pt]
Location &Mask ratio & NDS $\uparrow$ & mAP $\uparrow$ & mAVE $\downarrow$ \\ \midrule
\multirow{5}{*}{BEV} & 0.4 & 53.2 & 43.2 & 0.258 \\
    & \cellcolor{gray!40}0.5 & \cellcolor{gray!40}\textbf{53.5} & \cellcolor{gray!40}\textbf{42.7} & \cellcolor{gray!40}\textbf{0.255} \\
                    & 0.6 & 52.9 & 42.2 & 0.260 \\
                    & 0.75 & 53.4 & 42.0 & 0.258 \\
                    & 0.9 & 52.9 & 42.4 & 0.263 \\ \midrule
\multirow{3}{*}{BEV \& PV} & \cellcolor{gray!40}0.5 \& 0.5 & \cellcolor{gray!40}\textbf{53.7} & \cellcolor{gray!40}\textbf{43.0} & \cellcolor{gray!40}\textbf{0.249} \\
                        & 0.5 \& 0.65 & 53.6 & 42.6 & 0.260 \\
                        & 0.5 \& 0.75 & 53.3 & 42.6 & 0.258 \\
                    
\bottomrule[1pt]
\end{tabular}
}
\caption{Ablation of the mask ratio.}
\label{tab3}
\end{table}

\begin{table}[t]
\centering
\resizebox{\linewidth}{!}{
\begin{tabular}{c|c|ccccc}
\toprule[1pt]
\multirow{4}{*}{BEV}  & $\alpha_1$ & $1e^{-5}$  & $2e^{-5}$ & $5e^{-5}$ & $1e^{-4}$& $2e^{-4}$ \\ \cmidrule{2-7}
 & NDS $\uparrow$ & 53.3 & 53.2 & \textbf{53.5} & 53.3 & 52.7 \\ \cmidrule{2-7}
    & mAP $\uparrow$ & 42.0 & 42.3 & 42.7 & \textbf{42.9} & 42.0 \\ \cmidrule{2-7}
 & mAVE $\downarrow$ & \textbf{0.255} & 0.258 & \textbf{0.255} & 0.259 & 0.259 \\ \midrule
\multirow{4}{*}{BEV \& PV}  & $\alpha_2$ & $2e^{-4}$  & $5e^{-4}$ & $1e^{-3}$ & $2e^{-3}$ & $4e^{-3}$ \\ \cmidrule{2-7}
 & NDS $\uparrow$ & \textbf{53.9} & 53.1 & 53.7 & 53.0 & 53.6 \\ \cmidrule{2-7}
    & mAP $\uparrow$ & 42.8 & 42.1 & \textbf{43.0} & 41.9 & 42.6 \\ \cmidrule{2-7}
 & mAVE $\downarrow$ & 0.253 & 0.257 & \textbf{0.249} & 0.260 & 0.259 \\
                    
\bottomrule[1pt]
\end{tabular}
}
\caption{Ablation study on the loss weights of temporal features reconstruction. In BEV \& PV, we fix the $\alpha_1$ to $5e^{-5}$.}
\vspace{-0.8em}
\label{tab4}
\end{table}

\begin{table}[t]
\centering
\resizebox{\linewidth}{!}{
\begin{tabular}{c|cc|c}
\toprule[1pt]
Masked Feature Type & NDS $\uparrow$ & mAP $\uparrow$ & mAVE $\downarrow$ \\ \midrule
 -  & 52.9 & 41.6 & 0.263 \\
 spatial  & 53.1 & 42.3 & 0.254 \\
 temporal & 53.2 & 42.2 & 0.263 \\
 \rowcolor{gray!40}
 spatial + temporal & \textbf{53.7} & \textbf{43.0} & \textbf{0.249} \\
\bottomrule[1pt]
\end{tabular}
}
\caption{Ablation of the mask feature type on PV.}
\label{tab5}
\end{table}

\begin{table}[t]
\centering
\resizebox{\linewidth}{!}{
\begin{tabular}{c|cc|c}
\toprule[1pt]
Frames Num. & mAP $\uparrow$ & mAVE $\downarrow$ & FPS \\ \midrule
 2  & 39.0 (+0.9)  & 0.296 (-1.4\%) & 28.3 \\
\rowcolor{gray!40}
 4  & 43.2 (+1.6) & 0.253 (-1.0\%) & 26.1 \\
 8 & 45.1 (+0.4) & 0.238 (-1.8\%) & 20.2 \\

\bottomrule[1pt]
\end{tabular}
}
\caption{Ablation of the number of input frames.}
\label{tab6}
\end{table}

\subsection{Ablation Study} 
We conduct ablation study on the backbone ResNet101 (teacher) and ResNet50 (student) with $T_{tea}=8$ and $T_{stu}=4$, respectively, unless otherwise stated.

\textbf{Effects of loss components.} In Table \ref{tab2}, we evaluate the effects of each loss component on NDS and mAP. Notably, employing temporal feature reconstruction solely on BEV yields higher accuracy (NDS: $\uparrow 0.6$, mAP: $\uparrow 1.1$) than the other two loss terms. Moreover, the other two loss terms can also improve 0.7 mAP. When applying temporal feature reconstruction to both BEV and PV, we observe significant enhancements in NDS and mAP, reaching 53.7 and 43.0, respectively. This emphasizes the efficacy of temporal reconstruction in providing models with long-term memory object detection. In addition to the features in the encoder layers, we also focus on the decoded features. They encapsulate high-level spatio-temporal representation. As a result, transferring knowledge from the decoded features can also bring certain performance gains. 

\begin{figure*}[ht]
\begin{center}
\includegraphics[width=\linewidth]{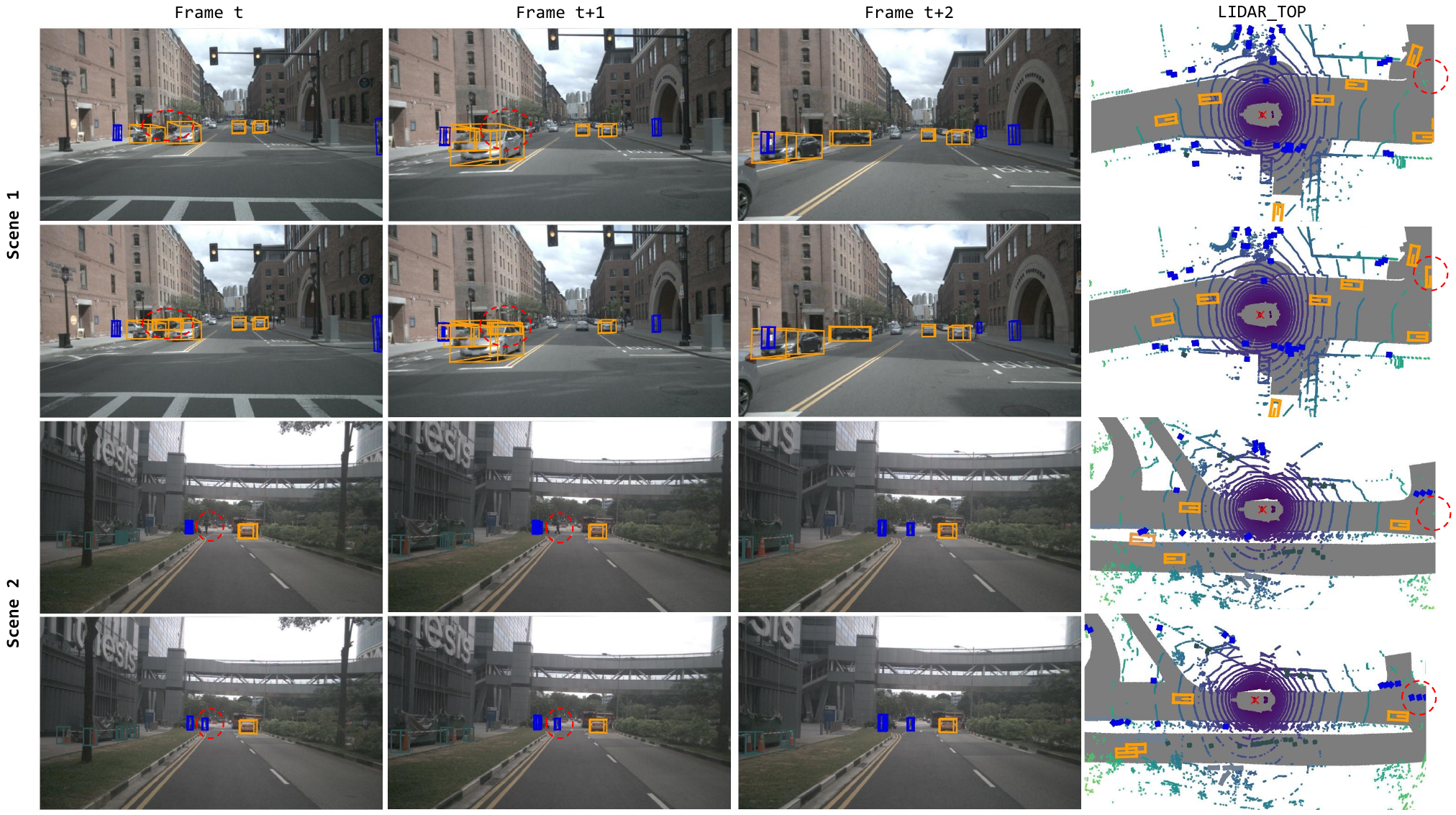}
\end{center}
\caption{Qualitative results over three consecutive frames (front camera) in two scenes. The first and third row show the prediction made by the baseline model, while the second and fourth row demonstrate the predictive results of TempDistiller. In the last column, the LiDAR point cloud in BEV is display for frame $t+1$, except the last row (for $t+2$) due to the limited BEV distance. TempDistiller successfully predicts an occluded car merging into the main road and a pedestrian crossing the street in the distance, highlighted by red dotted circles.}
\label{fig:visualization}
\end{figure*}

\textbf{Impacts of mask ratio.} We investigate various mask ratios for temporal feature reconstruction in both BEV and PV. As shown in Table \ref{tab3}, we can identify the optimal mask ratio to be 0.5 for both BEV and PV. The core idea of masked feature reconstruction is utilizing the residual features to reconstruct complete feature maps. A high mask ratio results in a poor representation of residual features, while a low mask ratio simplifies the generator's learning process, enabling shortcuts that lead to local optima. Furthermore, a low mAVE demonstrates that temporal reconstruction is conducive to focusing on dynamic objects and estimating their velocity.

\textbf{Effects of loss weights.} We examine the effects of loss weights in Table \ref{tab4}, which is significant to hybrid loss optimization. We find that $\alpha_1=5e^{-5}$ yields optimal results with the best NDS and mAVE, as well as relatively high mAP. Then we fix $\alpha_1$ to $5e^{-5}$ and tune $\alpha_2$. When $\alpha_2=1e^{-3}$, we can achieve 43.0 mAP and the lowest error of velocity estimation, and also obtain a relatively high NDS.

\textbf{Masked feature type.} We verify different feature types for reconstruction on PV in Table \ref{tab5}. Reconstructing either spatial features or temporal features can slightly boost performance (spatial: $\uparrow$ 0.2 NDS and $\uparrow$ 0.7 mAP, temporal: $\uparrow$ 0.3 NDS and $\uparrow$ 0.6 mAP). Notbaly, combining these two features types results in further performance gains through the feature reconstruction.

\textbf{The number of input frames.} We evaluate various number of input frames in Table \ref{tab6}. We find the student model with 4 frames can strike a balance between accuracy and efficiency.

\subsection{Visualization} 
We provide qualitative results to visualize the prediction of the model with and without TempDistiller, which highlights the superior performance of the proposed TempDistiller. Specifically, TempDistiller exhibits enhanced capabilities in detecting occluded objects and distant targets. As shown in the second row in Fig. \ref{fig:visualization}, an occluded vehicle about to merge onto the main road can be detected earlier. Furthermore, TempDistiller successfully identifies a distant pedestrian in advance (see the last row in Fig. \ref{fig:visualization}). These instances substantiate the efficacy of the proposed method in achieving comparable results using fewer frames compared to methods dealing with longer temporal sequences.

\section{Conclusions}
In this paper, we propose TempDistiller, a temporal knowledge distillation method for reconstructing long-term temporal features and exploring temporal relational knowledge. Our method involves aggregating long-term temporal features from the teacher as a reconstruction objective. Subsequently, we leverage this objective to reconstruct the masked student features. This design allows TempDistiller to effectively capture long-term temporal knowledge even with a reduced number of input frames. In addition, temporal relational distillation can improve the detection of dynamic objects. Experiments show that our approach can strike a balance between accuracy and efficiency, with significant improvement in velocity estimation.

\textbf{Limitations and future works.} Constrained by the parallel temporal fusion paradigm, the speed of TempDistiller is still affected by the number of input frames. Moreover, the number of input frames is still limited. Incorporating more frames is challenging. In the future, we will explore temporal knowledge with more frames through sequential temporal fusion, aiming to solve the problems of limited speed and temporal knowledge forgetting.

\clearpage
{
    \small
    \bibliographystyle{ieeenat_fullname}
    \bibliography{main}
}


\end{document}